%% file: main.tex
\documentclass[runningheads]{llncs}

\usepackage{eccv}

\usepackage{eccvabbrv}

\usepackage{epsfig}
\usepackage{graphicx}
\usepackage{amsmath}
\usepackage{amssymb}
\usepackage{paralist}
\usepackage{float}
\usepackage{stfloats}
\usepackage{booktabs}
\usepackage{multicol}
\usepackage{multirow}
\usepackage{bm}
\usepackage{bbm}
\usepackage{diagbox}
\usepackage{makecell}
\usepackage{algorithm}  
\usepackage{algorithmic} 
\usepackage{color, colortbl}
\newcommand{\gr}{\rowcolor[gray]{.95}} 

\usepackage[accsupp]{axessibility}  

\usepackage{wrapfig}
\usepackage[misc]{ifsym}

\usepackage{hyperref}

\usepackage{orcidlink}

\def\ours{HENet}
\newcommand{\figref}[1]{Figure~\ref{#1}}%
\newcommand{\tabref}[1]{Table~\ref{#1}}%
\newcommand{\secref}[1]{Section~\ref{#1}}
\renewcommand{\eqref}[1]{Eq.~(\ref{#1})}

\newcommand{\blue}[1]{{\color{blue}{#1}}}
\newcommand{\red}[1]{{\color{red}{#1}}}

\begin{document}

\title{HENet: Hybrid Encoding for End-to-end Multi-task 3D Perception from Multi-view Cameras} 

\titlerunning{HENet}

\author{Zhongyu Xia\inst{1} \quad
Zhiwei Lin\inst{1} \quad
Xinhao Wang\inst{1} \quad
Yongtao Wang\inst{1}\textsuperscript{(\Letter)} \\
Yun Xing\inst{2} \quad
Shengxiang Qi\inst{2} \quad
Nan Dong\inst{2} \quad
Ming-Hsuan Yang\inst{3}
}

\authorrunning{Z. Xia et al.}

\institute{Wangxuan Institute of Computer Technology, Peking University 
\and
Chongqing Changan Automobile Co., Ltd. \and University of California, Merced
\\
\email{\{xiazhongyu,zwlin,wangxinhao,wyt\}@pku.edu.cn,\\ dongnan1@changan.com.cn, mhyang@ucmerced.edu}
}

\maketitle

\input{sec/0_abstract}    
\input{sec/1_intro}

\input{sec/2_related}

\input{sec/3_method}
\input{sec/4_exp}

\input{sec/5_conclusion}


\section*{Acknowledgments}

This work was supported by National Key R\&D Program of China (Grant No. 2022ZD0160305).

%
%
\bibliographystyle{splncs04}
\bibliography{main}
\end{document}

%% file: sec/0_abstract.tex
\begin{abstract}

Three-dimensional perception from multi-view cameras is a crucial component in autonomous driving systems, which involves multiple tasks like 3D object detection and bird's-eye-view (BEV) semantic segmentation. 
To improve perception precision, large image encoders, high-resolution images, and long-term temporal inputs have been adopted in recent 3D perception models, bringing remarkable performance gains. 
However, these techniques are often incompatible in training and inference scenarios due to computational resource constraints. 
Besides, modern autonomous driving systems prefer to adopt an end-to-end framework for multi-task 3D perception, which can simplify the overall system architecture and reduce the implementation complexity.
However, conflict between tasks often arises when optimizing multiple tasks jointly within an end-to-end 3D perception model. 
To alleviate these issues, we present an end-to-end framework named HENet for multi-task 3D perception in this paper. 
Specifically, we propose a hybrid image encoding network, using a large image encoder for short-term frames and a small image encoder for long-term temporal frames. 
Then, we introduce a temporal feature integration module based on the attention mechanism to fuse the features of different frames extracted by the two aforementioned hybrid image encoders. 
Finally, according to the characteristics of each perception task, we utilize BEV features of different grid sizes, independent BEV encoders, and task decoders for different tasks. 
Experimental results show that HENet achieves state-of-the-art end-to-end multi-task 3D perception results on the nuScenes benchmark, including 3D object detection and BEV semantic segmentation. 
%

\end{abstract}

%% file: sec/1_intro.tex
\section{Introduction}
\label{sec:intro}

Efficient and accurate perception of the surrounding environment from multi-view cameras is essential for the autonomous driving system, which serves as the basis for subsequent trajectory prediction and motion planning tasks.
A desirable 3D perception system should handle several tasks simultaneously, including 3D
\begin{wrapfigure}[14]{r}{0.5\textwidth}
    \setlength{\abovecaptionskip}{-0.cm}
    \centering
    \includegraphics[width=\linewidth]{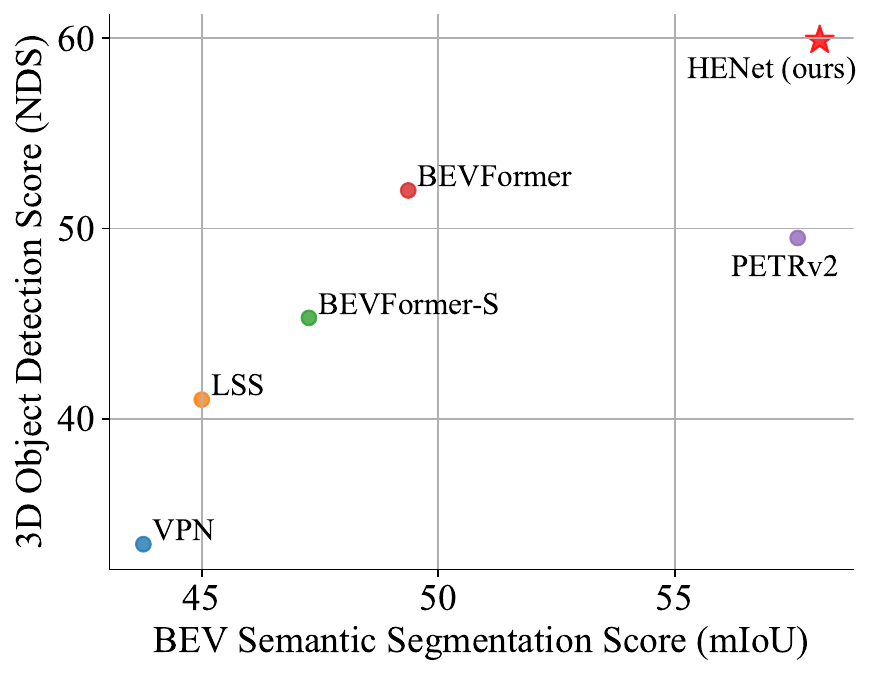}
    \caption{\textbf{Comparisons of end-to-end multi-task results on nuScenes {\tt{val}} set.}
    }
    \label{fig:multitask_result}
\end{wrapfigure}
object detection and bird's-eye-view (BEV) semantic segmentation.
There has been an increasing emphasis on end-to-end multi-task frameworks,  
as such systems have the potential to streamline the overall architecture and alleviate implementation complexities.

However, end-to-end multi-task 3D perception faces the following challenges.
First, when designing high-performance camera-based 3D perception models, researchers usually exploit techniques like higher-resolution images, longer temporal inputs, and larger image encoders to improve 3D perception accuracy.
Nevertheless, \textit{simultaneously employing these techniques on a single perception model would result in a formidable cost during training}.
To alleviate this issue, a few works \cite{SOLOFusion, 3d-man} store past information in memory, which has disadvantages such as inconsistency of temporal features and inefficiency of data augmentations. 
Therefore, many of the latest methods \cite{bevformerv2, StreamPetr, SparseBEV, HoP} do not adopt this strategy, but recompute features of past frames, albeit at the expense of training costs.

Second, to process long-term temporal inputs, many works~\cite{BEVDepth, BEVStereo, BEVDet4D} directly sum or concatenate the features from different frames in BEV along the channel dimension, showing unsatisfactory perception performance with longer time sequence.
The reason is that \textit{the features of the moving objects are misaligned and scattered across large regions in BEV along their trajectories in different frames}.
Therefore, it is necessary to introduce dynamic alignment mechanisms~\cite{PETRv2, StreamPetr} for location correction of moving objects.

Third, for end-to-end multi-task learning, existing works~\cite{PETRv2, UniAD, BEVFormer} use a shared encoding network with multiple decoders for different tasks. 
However, experimental results in these works show that \textit{jointly learning multiple tasks in the end-to-end fashion is often sub-optimal}, \textit{i.e.}, the performance of each task in multi-task learning is lower than standing-alone training.
To alleviate this issue, some works~\cite{PETRv2} propose adjusting the loss weight for each task, but there is no comprehensive analysis of why a conflict exists between the tasks.

In this paper, we present \ours, an end-to-end multi-task 3D perception framework for multi-view cameras. 
To integrate large image encoders, high-resolution images, and long-term inputs, we propose a hybrid image encoding network that adopts different resolutions and image encoders for different frames.
Specifically, we use high-resolution inputs, a large image backbone, and a complex perspective transformation network for short-term frames to produce high-precision BEV features.
For long-term frames, low-resolution inputs are chosen, and a small image backbone and a simple perspective transformation network are employed to generate BEV features efficiently.
The proposed hybrid image encoding network can be easily incorporated into existing perception models.
Then, we introduce a temporal integration module to align and fuse the BEV features from multi-frames dynamically. 
Specifically, in this module, we present a temporal backward and forward process with adjacent frame fusion modules (AFFM) to aggregate BEV features, addressing the problem of aligning moving objects with the attention mechanism.
Finally, we deeply analyze the conflict between 3D object detection and BEV semantic segmentation in multi-task learning and find the critical problem is that different tasks prefer different BEV feature grid sizes.
Based on this observation, we select BEV features of different grid sizes for different tasks.
%
The selected features are sent to independent BEV encoding networks and task decoders to alleviate the task conflicts further to obtain the final 3D perception results.

The main contributions of this work are summarized as follows:
\begin{compactitem}
    \item
    We present an end-to-end multi-task 3D perception framework with a hybrid image encoding network to take advantage of high-resolution images, long-term inputs, and large image encoders with a smaller training cost.
    
    \item
    We introduce a temporal integration module based on the attention mechanism to fuse the multi-frame BEV features and enable the dynamic inter-frame alignment for moving objects.

    \item 
    We analyze the task conflicts in end-to-end multi-task learning and propose a feature size selection and independent feature encoding to alleviate this problem.
    \item 
    We achieve state-of-the-art results in end-to-end multi-task learning on nuScenes, including 3D object detection and BEV semantic segmentation tasks.
    
\end{compactitem}

%% file: sec/2_related.tex
\section{Related Work}
\label{sec:related}

\subsection{Multi-View 3D Object Detection}

Three-dimensional object detection from images is a crucial task in multi-task 3D perception.
Early approaches focus on predicting objects directly from monocular images~\cite{wang2019pseudo,Fcos3d,D4LCN,M3d-rpn,DFM,dd3d,CaDNN,OFT}. 
Recently, multi-view cameras become the default sensors for autonomous driving cars, which can provide more information.
Based on the type of view transformations, current multi-view 3D object detection works can be divided into two categories, \textit{i.e.}, BEV-based methods~\cite{BEVDet,BEVDepth,BEVDet4D,BEVStereo,SOLOFusion,aedet,fastBEV,polarformer,BEVFormer,bevformerv2,sts,HoP} and sparse query methods~\cite{simMOD,DETR3D,PETR,PETRv2,3d-man,sparse4d,sparse4dv2,StreamPetr,SparseBEV,far3d}.

\noindent\textbf{BEV-based Methods}.
BEVDet~\cite{BEVDet} utilizes Lift-Splat-Shoot (LSS)~\cite{LSS} to generate the BEV feature from multi-view image features with depth prediction.
To address the issue of inaccurate depth estimation in BEVDet~\cite{BEVDet}, BEVDepth~\cite{BEVDepth} introduces extrinsic parameters for the depth prediction network and adds additional depth supervision from LiDAR point clouds. 
BEVDet4D~\cite{BEVDet4D} incorporates temporal information by aligning BEV features from different frames through ego-vehicle transformations.
Based on BEVDepth~\cite{BEVDepth}, BEVStereo~\cite{BEVStereo} and STS~\cite{sts} present temporal stereo techniques to improve depth prediction further. 
%
%
To gather long-term temporal information, BEVFormer~\cite{BEVFormer} and Polarformer~\cite{polarformer} regard the BEV feature as queries and utilize cross-attention to aggregate information from all past frames. 
%
%
BEVFormerv2~\cite{bevformerv2} further introduces perspective space supervision to enhance the performance of BEVFormer.
%
SOLOFusion \cite{SOLOFusion} proposes to first fuse adjacent short-term series BEV features and then fuse long-term series.
HoP~\cite{HoP} designs a historical object prediction module, which can be easily plugged into any temporal 3D object detectors.
%
AeDet~\cite{aedet} proposes azimuth-equivariant convolution operations and anchors to unify the BEV representation in different azimuths.
%
%

\noindent\textbf{Sparse Query Methods}.
As a pioneer in sparse query method, DETR3D~\cite{DETR3D} extends DETR~\cite{DETR} by utilizing a sparse 3D object query set to index features.
3D-MAN~\cite{3d-man} designs an alignment and aggregation module to extract temporal features from the memory bank that stores information generated by a single frame detector.
PETR~\cite{PETR} builds upon DETR3D~\cite{DETR3D} by aggregating image features with 3D position information.
PETRv2~\cite{PETRv2} introduces temporal information into 3D position embedding in PETR~\cite{PETR} to allow temporal alignment for object positions across different frames. 
Sparse4D~\cite{sparse4d} assigns and projects 4D keypoints to generate different view, scale, and timestamp features.
Sparse4Dv2~\cite{sparse4dv2} improves the temporal fusion module to reduce the computational complexity and enable long-term fusion.
StreamPetr~\cite{StreamPetr} presents an efficient intermediate representation to transfer temporal information like BEV-based methods to avoid repeated calculation of features. 
SparseBEV~\cite{SparseBEV} designs a scale-adaptive self-attention module for query feature interaction and proposes spatiotemporal sampling and adaptive mixing to aggregate temporal features into current queries.
To explore long-range detection, Far3D~\cite{far3d} uses a perspective-aware aggregation module to capture features of long-range objects and designs a denoising method to improve query propagation.

\subsection{BEV Semantic Segmentation}
Another task in multi-task 3D perception involves BEV semantic segmentation, which aims to comprehend the overall scene.
Many works~\cite{LSS,fiery,yang2021projecting,roddick2020predicting,VPN} share a common paradigm like BEVDet~\cite{BEVDet}, except for different task heads.
VPN~\cite{VPN} trains their network in a 3D graphics environment and utilizes domain adaptation techniques to transfer it for handling real-world data.
M2BEV~\cite{xie2022m} adeptly transforms multi-view 2D image features into 3D BEV features in the ego vehicle coordinates. This BEV representation is paramount, as it facilitates sharing a singular encoder across diverse tasks.
CVT~\cite{zhou2022cross} employs a cross-view attention mechanism to implicitly learn the mapping from an individual camera view to a canonical map view representation. Each camera utilizes position embeddings dependent on its intrinsic and extrinsic calibration. 
HDMapNet~\cite{li2022hdmapnet} encodes image features from surrounding cameras or point clouds from LiDAR and predicts vectorized map elements in a BEV map.

\subsection{End-to-end Multi-task Learning}

In autonomous driving scenarios, multi-task learning, usually including 3D object detection and BEV segmentation~\cite{PMF,OCNet,roddick2020predicting,VPN}, is crucial for scene understanding.
Current works often follow the joint training strategy of BEVFormer~\cite{BEVFormer}, which generates a unified BEV feature map for detection and segmentation tasks.
PETRv2~\cite{PETRv2} initializes two query sets for detection and segmentation tasks and sends each query set to corresponding task heads. 
The latest work UniAD~\cite{UniAD} integrates three major tasks, including perception, prediction, and planning, and six sub-tasks into an end-to-end framework with query design as the connection between tasks.

These works show conflicts between tasks when combining detection and segmentation tasks. 
Nevertheless, no comprehensive analysis has been reported to address this issue. 
In this work, we make an attempt to investigate the problem of performance degradation in multi-task training of 3D object detection and BEV semantic segmentation.
Most importantly, we propose an efficient multi-task 3D perception framework and an effective training strategy to alleviate this issue. 
Our model can achieve state-of-the-art 3D object detection and BEV segmentation results in multi-task settings.

%% file: sec/3_method.tex
\section{Method}
\label{sec:method}

\begin{figure*}[tp]
    \setlength{\abovecaptionskip}{-0.cm}
    \centering
    \includegraphics[width=0.97\linewidth]{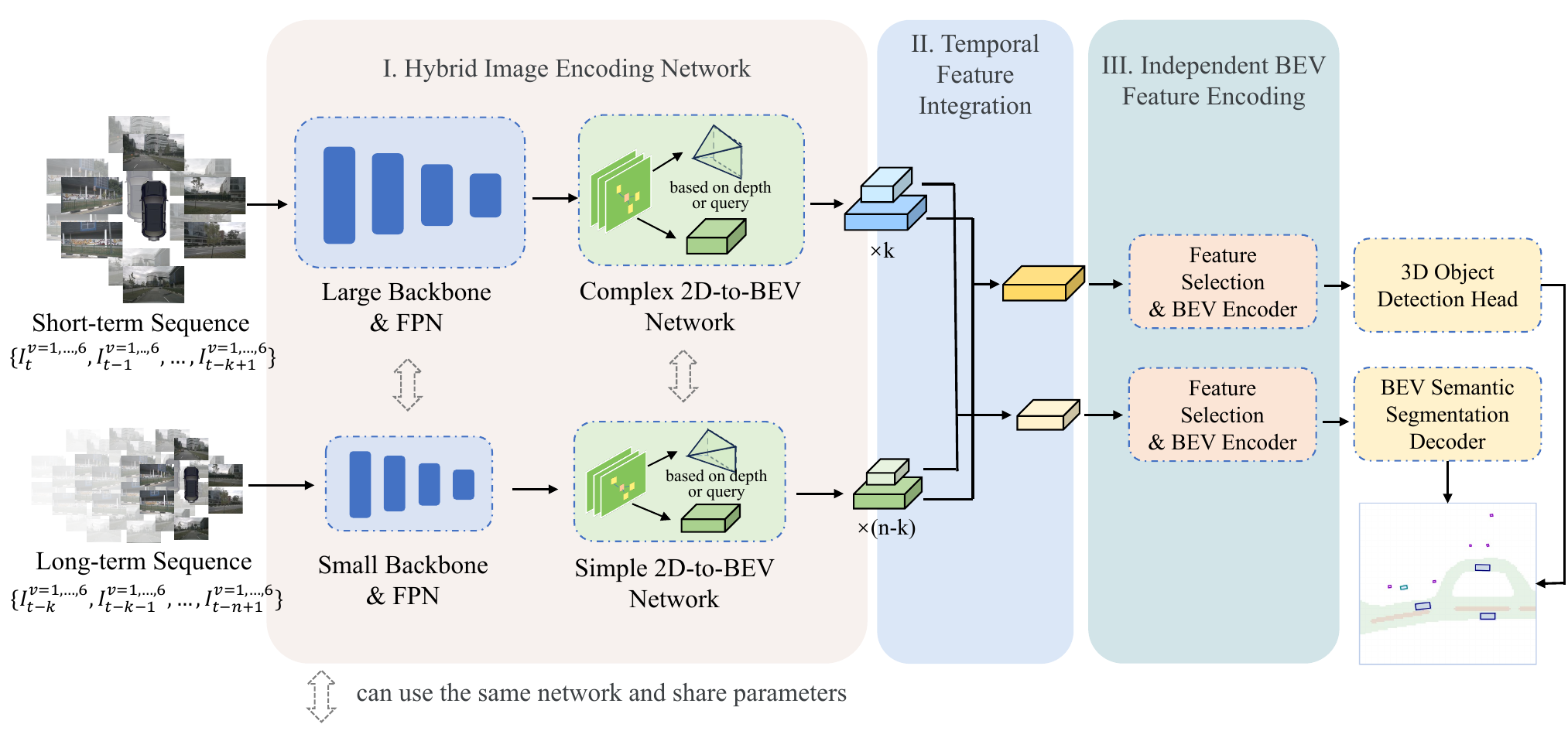}

    \caption{\textbf{Overall architecture of \ours.} i) Hybrid Image Encoding Network uses image encoders of varying complexity to encode long-sequence frames and short-term images, respectively. ii) Temporal Feature Integration module based on the attention mechanism fuses multi-frame features from multiple image encoders. iii) Based on the characteristics of different tasks, we select BEV feature maps of appropriate sizes and perform independent BEV encoding for each task.}
    \label{fig:arch}

\end{figure*}

We illustrate the overall architecture of \ours~in~\figref{fig:arch}.
Given temporal multi-view image inputs, a hybrid image encoding network extracts their BEV features.
Then, we use a temporal feature integration module to aggregate the multi-frame BEV features.
Finally, we send BEV features in different grid sizes to independent BEV feature encoders and decoders for different tasks to obtain the multi-task perception results.

\subsection{Hybrid Image Encoding Network}
As shown in~\figref{fig:arch}, the hybrid image encoding network contains two image encoders with different complexity. 
Specifically, the first one processes short-term frames, which scales up the high-resolution inputs and sends them to an image backbone and a feature pyramid network (FPN). 
Then, a complex 2D-to-BEV network is applied to produce high-precision BEV features.
The second image encoder handles the long-term sequences by down-sampling the inputs to low resolution and utilizing a small backbone and FPN to extract image features.
Similarly, a simple 2D-to-BEV network is adopted afterward for efficient BEV feature generation.
Some parts of the two image encoders can be shared. 
For example, we can use the same backbone with a different 2D-to-BEV network. 
However, the first image encoder is more complex than the second one.

The aforementioned hybrid image encoding network can be incorporated into existing multi-view 3D object detection methods.
Without loss of generality, we choose two popular BEV-based approaches, BEVDepth~\cite{BEVDepth} and BEVStereo~\cite{BEVStereo}, as examples to show how the proposed hybrid image encoding network works.
Specifically, for the first image encoder, which processes the short-term frames, we keep a high resolution for images and utilize a large backbone (\textit{e.g.}, VoVNetV2-99~\cite{v299}) followed by FPN~\cite{FPN} to extract the features. 
Then, multiple convolution layers and EM algorithm in BEVStereo~\cite{BEVStereo} are applied to generate stereo depth and frustum features. 
BEVPoolv2~\cite{bevpoolv2} converts frustum features into multi-scale BEV features.
As for the second image encoder, we first down-sample the image size for the long-term image sequences and then use a small backbone (\textit{e.g.}, ResNet-50) followed by FPN~\cite{FPN} to extract the multi-view image features. 
After that, we adopt the simple monocular depth estimation network in BEVDepth~\cite{BEVDepth} to obtain frustum features. 
%
%
Further, we use BEVPoolv2~\cite{bevpoolv2} to transform frustum features in perspective view into multi-scale BEV features according to the camera's intrinsic and extrinsic parameters. 
%
We select different BEV sizes for 3D object detection and BEV semantic segmentation tasks according to experiments in~\secref{sec:analysis}.
Specifically, we utilize BEV sizes of 256x256 for the 3D object detection task and 128x128 for the BEV semantic segmentation task.

\begin{figure}[t]
    \setlength{\abovecaptionskip}{-0.cm}
    \centering
    \includegraphics[width=0.8\linewidth]{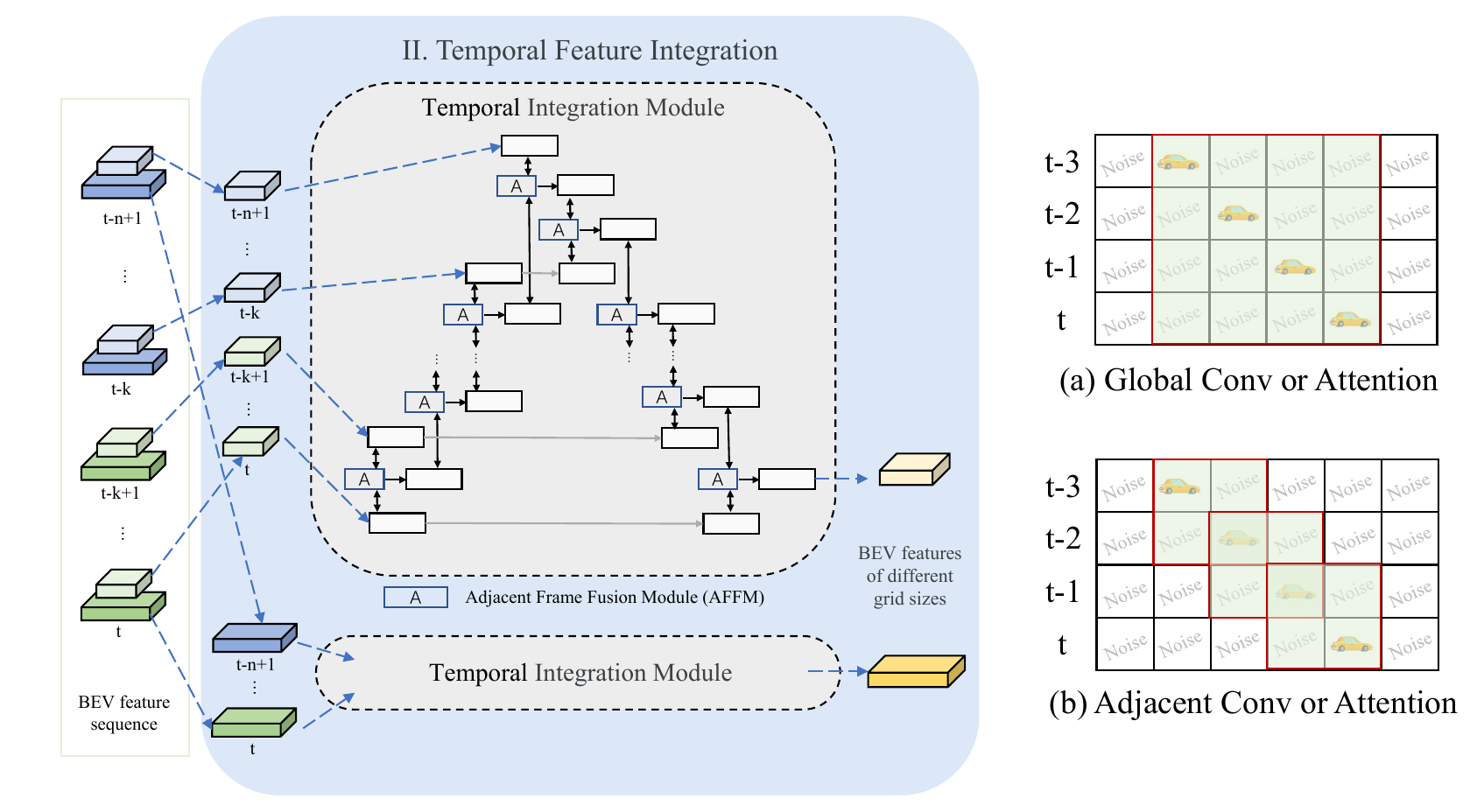}
    \caption{\textbf{Architecture of Temporal Feature Integration module.} We propose the adjacent frame fusion module (AFFM) and adopt the temporal fusion strategy with temporal backward and forward processes.}
    \label{fig:temporal}

\end{figure}

\subsection{Temporal Feature Integration}
\label{sec:temporal}

\begin{algorithm}[t]
\caption{Pseudo-code for \secref{sec:temporal}}
\label{algorithm1}
\begin{algorithmic}
\footnotesize 
\STATE \textbf{Input:} A series of BEV features $\{f_{-n+1}, ... , f_{-1}, f_0\}$. 
$f_0$ represents BEV feature of the current frame and $f_{-i}$ corresponds to the $i^{th}$ frame before $f_0$.
    
\FOR{i \textbf{from} 0 \textbf{to} n-2}
\STATE $f_{-(i+1)} \leftarrow \text{AFFM}(f_{-i}, f_{-(i+1)})$
\ENDFOR

\FOR{i \textbf{from} n-2 \textbf{to} 0}
\STATE $f_{-i} \leftarrow \text{AFFM}(f_{-(i+1)}, f_{-i})$
\ENDFOR

\textbf{return} $f_{0}$

\end{algorithmic}
\end{algorithm}

After generating multi-frame BEV features by the hybrid image encoding network, we adopt temporal integration modules to fuse the BEV features, as shown in~\figref{fig:temporal}.
The temporal integration module consists of a backward process and a forward process.
The backward process fuses features of the current frame to past frames, while the forward process aggregates features from the past to the current frame.
Algorithm~\ref{algorithm1} provides the pseudo-code of the aforementioned temporal integration process.
In each process step, we employ an adjacent frame fusion module (AFFM) with a cross-attention to fuse the BEV features from two adjacent frames.
Specifically, assuming the BEV features of two frames are $f_i$ and $f_j$, AFFM can be formulated as:
\begin{equation}
\text{AFFM}(  f_i, f_j) =  f_j + \gamma \times  \text{Avg}( 
     Atn(\langle f_i, f_j \rangle, f_i, f_i), 
     Atn(\langle f_i, f_j \rangle, f_j, f_j)),
\end{equation}
where $Avg(\cdot)$ represents the average operator, $\gamma$ is a learnable scaling parameter, $\langle \cdot, \cdot \rangle$ denotes concatenation, and $Atn(\cdot,\cdot,\cdot)$ is the attention module:
\begin{equation}
Atn(q,k,v) = softmax(\frac{qk^\top}{\sqrt{d}})v.
\end{equation}
In the backward process, $j=i-1$, while in the forward process, $j=i+1$.

As shown in~\figref{fig:temporal}, using adjacent attention will introduce less noise than directly applying global attention or convolution across all frames.
%
With adjacent attention, AFFM can align the features of moving objects and avoid fusing redundant background information.
Besides, the attention mechanism allows us to adopt AFFM to both BEV-based and sparse query methods.
%

\subsection{Independent BEV Feature Encoding}

\begin{figure}[t]
    \setlength{\abovecaptionskip}{-0.cm}
    \centering
    \includegraphics[width=0.75\linewidth]{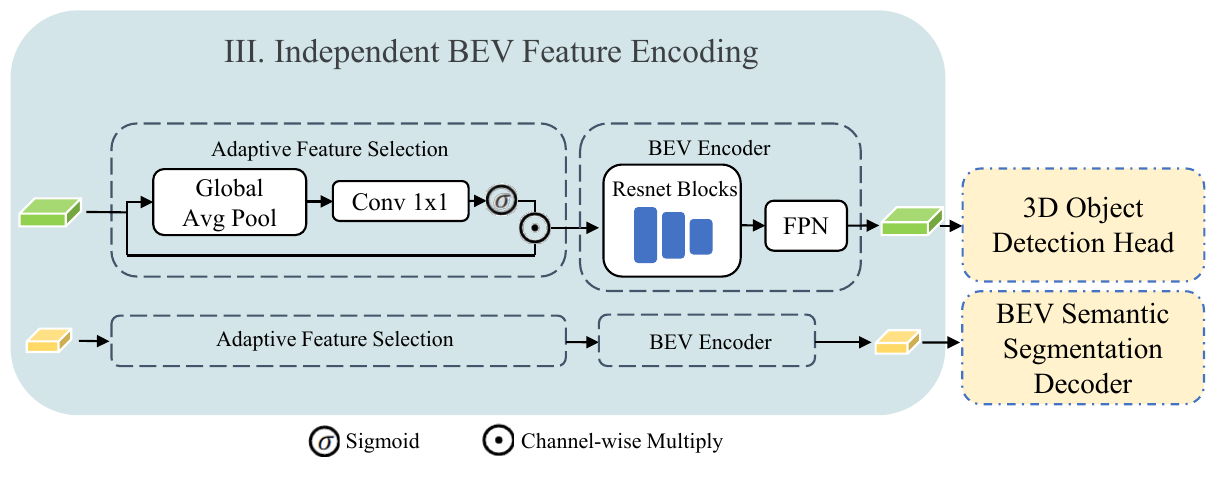}
    \caption{\textbf{Design of Independent BEV Feature Encoding.} Each task decoder is provided with BEV feature maps in different grid sizes through independent adaptive feature selection and BEV encoding.}
    \label{fig:taskencoder}

\end{figure}

Based on our empirical analysis in \secref{sec:analysis}, different tasks prefer different sizes of BEV features.
Thus, after obtaining the fused multi-scale BEV features, we first assign different sizes of BEV features to other tasks.
We then encode the BEV feature for each task independently.
Inspired by BEVFusion~\cite{bevfusion}, the proposed encoding process comprises adaptive feature selection and BEV encoding.

Specifically, the adaptive feature selection $f_{adaptive}(\cdot)$ applies a simple channel attention module to select important features as follows:
\begin{equation}
f_\text{adaptive}(\mathbf{F})=\sigma\left(\mathbf{W}f_\text{avg}(\mathbf{F})\right) \cdot \mathbf{F},
\end{equation}
where $\mathbf{F}\in \mathbb{R}^{X\times Y\times C}$ is the BEV features, $\mathbf{W}$ denotes linear transform matrix, $f_\text{avg}$ indicates the global average pooling, and $\sigma$ represents the Sigmoid function.
For the BEV encoder, we adopt three ResNet~\cite{resnet} residual blocks and a simple FPN~\cite{FPN} following existing BEV-based methods~\cite{BEVDepth,BEVDet4D} to perform local feature integration on the BEV feature map. 
Notably, the adaptive feature selection and BEV encoders for different tasks share the same architecture but have different weights.

\subsection{Decoders and Losses}

We adopt Centerpoint~\cite{centerpoint} as the 3D object detection decoder and SegNet~\cite{segnet} as the BEV semantic segmentation decoder. 
The classification and regression loss of Centerpoint~\cite{centerpoint} is represented by $\mathcal{L}_{cls}$ and $\mathcal{L}_{bbox}$. 
We use focal loss~\cite{focalloss} for the BEV semantic segmentation task. 
The loss functions for vehicles, drivable areas, and lane lines are denoted as $\mathcal{L}_{segveh}$, $\mathcal{L}_{segarea}$, and $\mathcal{L}_{segdiv}$, respectively.

The final loss function is defined as 
\begin{equation}
\begin{split}
\mathcal{L} = &\alpha_{depth} \mathcal{L}_{depth} + \alpha_{cls} \mathcal{L}_{cls} + \alpha_{bbox} \mathcal{L}_{bbox} + \\
&\alpha_{segveh} \mathcal{L}_{segveh} + \alpha_{segarea} \mathcal{L}_{segarea} + \alpha_{segdiv} \mathcal{L}_{segdiv},
\end{split}
\end{equation}
where $\mathcal{L}_{depth}$ is a binary cross entropy loss for depth estimation and $\alpha$ is the loss weight.

%% file: sec/4_exp.tex
\section{Experiments}
\label{sec:exp}

\subsection{Implementation Details}


We train \ours~in end-to-end manner for multi-tasks, including 3D object detection and BEV semantic segmentation in the same way as LSS~\cite{LSS}.
We choose VovNet-99~\cite{v299, dd3d} with $896\times 1600$ image resolution for the large image encoder and select ResNet-50~\cite{resnet} with $256\times 704$ image resolution for the small image encoder, respectively.
%
For the input temporal sequence, we set short-term frame number $k=2$ and long-term frame number $n=9$.
The weights of the hybrid image encoding network are initialized from pre-trained 3D detectors.
As analyzed in Sec.\ref{sec:analysis}, we choose BEV grid sizes of 0.4m (256$\times$256 BEV size) and 0.8m (128$\times$128 BEV size) for 3D object detection and BEV semantic segmentation, respectively.
The end-to-end multi-task models are trained for 60 epochs without CBGS.
%
Besides, to further compare HENet with some single-task methods, we train the single 3D object detection models of HENet for 12 epochs with CBGS~\cite{CBGS}.

\begin{table*}[t]
\setlength{\belowcaptionskip}{-0.cm}
\centering
\caption[results1]{\textbf{Comparison of end-to-end multi-task learning on the nuScences {\tt{val}} set}. 
\textbf{Time/E} represent the training time per epoch with FP32 on 8$\times$A800 GPUs.
$\textbf{mIoU}_{\textbf{v}}$, $\textbf{mIoU}_{\textbf{a}}$, and $\textbf{mIoU}_{\textbf{d}}$ represent the mIoU for vehicles, drivable area, and lane \& road divider, respectively. 
}

\label{table:exp1} 
\resizebox{\textwidth}{!}{
\begin{tabular}{l|ccc|c||cc|cccc}
\hline
{\textbf{Methods}} & \textbf{Backbone} & \textbf{Input Size} & \textbf{Frames} & \textbf{Time/E} & \textbf{NDS$\uparrow$} & \textbf{mAP$\uparrow$} & 
$\textbf{mIoU}\uparrow$ & 
$\textbf{mIoU}_{\textbf{v}}\uparrow$ & $\textbf{mIoU}_{\textbf{a}}\uparrow$ & $\textbf{mIoU}_{\textbf{d}}\uparrow$\\
\hline

\gr {VPN~\cite{VPN}} & ResNet50 & 900$\times$1600 & 1 & - & 33.4 & 25.7 & 43.8 & 37.3 & 76.0 & 18.0 \\
{LSS~\cite{LSS}} & ResNet50 & 900$\times$1600 &  1  & - & 41.0 & 34.4 & 45.0 & 42.8 & 73.9 & 18.3 \\
\gr {BEVFormer-S~\cite{BEVFormer}} & ResNet101-DCN & 900$\times$1600 & 1 & - & 45.3 & 38.0 & 47.3 & 44.4 & 77.6 & 19.8 \\
\hline

{BEVFormer~\cite{BEVFormer}} & ResNet101-DCN & 900$\times$1600 & 5 & 213min & 52.0 & 41.2 & 49.4 & 46.7 & 77.5 & 23.9 \\
\gr {UniAD~\cite{UniAD}} & ResNet101 & 900$\times$1600 & 6 & 253min & 49.9 & 38.2 & - & - & 69.1 & 25.7 \\
\ours~(Ours) & ResNet101 \& ResNet50 & 640$\times$1152\&256$\times$704 & 2 \& 3 & 60min & 56.4 & 47.1 & 53.9 & 44.5 & 77.0 & 40.1 \\
\hline

\gr {PETRv2~\cite{PETRv2}} & V2-99 & 640$\times$1600 &2 & 75min & 49.5 & 40.1 & 57.6 & 49.4 & 79.1 & \textbf{44.3} \\
\ours~(Ours) & V2-99 \& ResNet50 & 640$\times$1152 \& 256x704 & 2 & 58min & 58.0 & 48.7 & 56.9 & 47.6 & 80.2 & 42.8 \\

\gr {\ours~(Ours)} & V2-99 \& ResNet50 & 640$\times$1152 \& 256$\times$704 & 2 \& 7 & 71min & \textbf{59.9} & \textbf{49.9} & \textbf{58.0} & \textbf{49.5} & \textbf{81.3} & 43.4 \\

\hline
\end{tabular}
}
\end{table*}

\begin{figure}[t]
    \setlength{\abovecaptionskip}{-0.cm}
    \centering
    \includegraphics[width=0.95\linewidth]{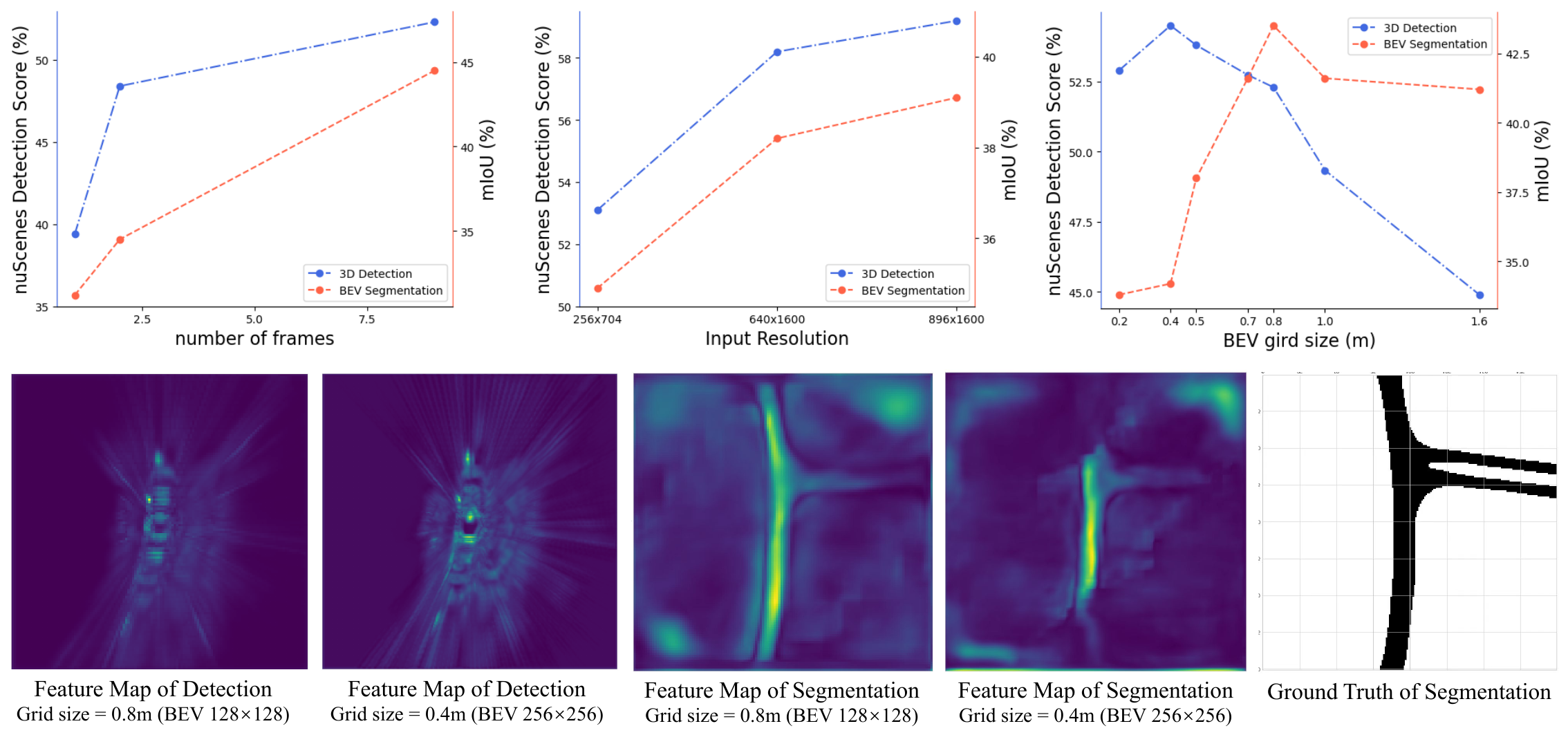}
    \caption{\textbf{Analysis of similarities and differences between 3D object detection and BEV semantic segmentation.} Experimental results show that each task has a suitable BEV grid size. The suitable grid size of BEV Semantic Segmentation is larger than 3D Object Detection.
    }
    \label{fig:diff}

\end{figure}

\subsection{Dataset and Metrics}
We evaluate our model on the nuScenes~\cite{nuScenes} dataset, a large-scale autonomous driving dataset containing 1,000 driving scenes (700 for training, 150 for validation, and 150 for testing), including cities, highways, and rural roads. 
%
%
Each scene contains various objects, such as vehicles, pedestrians, and bicycles. 


For 3D object detection evaluation, NuScenes provides a set of evaluation metrics, including mean Average Precision (mAP) and five true positive (TP) metrics: ATE, ASE, AOE, AVE, and AAE for measuring translation, scale, orientation, velocity, and attribute errors respectively. The overall performance is measured by the nuScenes Detection Score (NDS), which is the composite of the above metrics.
For BEV semantic segmentation, we use mean intersection over union (mIoU) as the metric following the settings of LSS~\cite{LSS}.

\begin{table*}[t]
\setlength{\belowcaptionskip}{-0.cm}
\centering
\caption[results2]{\textbf{Comparison of 3D object detection results on the nuScences {\tt{val}} set}. $*$ indicates the result is benefited from the perspective pre-training. $\dagger$ indicates using one temporal frame information. $\ddagger$ denotes integrating two or more temporal frames. The best and second best results are marked in \red{red} and \blue{blue}.}
\label{table:exp2} 
\resizebox{\textwidth}{!}{
\begin{tabular}{l|cc||cc|ccccc}
\hline
{\textbf{Methods}} & \textbf{Backbone} & \textbf{Input Size}  & \textbf{NDS$\uparrow$} & \textbf{mAP$\uparrow$}  & \textbf{mATE$\downarrow$} & \textbf{mASE$\downarrow$} & \textbf{mAOE$\downarrow$} & \textbf{mAVE$\downarrow$} & \textbf{mAAE$\downarrow$}\\
\hline
{BEVDet~\cite{BEVDet}} & ResNet50 & 256$\times$704 & 37.9 & 29.8 & 0.725 & 0.279 & 0.589 & 0.860 & 0.245 \\
\gr {BEVDet4D~\cite{BEVDet4D}$\dagger$}  & ResNet50 & 256$\times$704 & 45.7 & 32.2 & 0.703 & 0.278 & 0.495 & 0.354 & 0.206 \\
{PETRv2~\cite{PETRv2}$\dagger$}  & ResNet50 & 256$\times$704 & 45.6 & 34.9 & 0.700 & 0.275 & 0.580 & 0.437 & 0.187 \\
\gr {BEVStereo~\cite{BEVStereo}$\dagger$} & ResNet50 & 256$\times$704 & 50.0 & 37.2 & 0.598 & 0.270 & 0.438 & 0.367 & 0.190 \\
{SOLOFusion~\cite{SOLOFusion}$\ddagger$} & ResNet50 & 256$\times$704 & 53.4 & 42.7 & 0.567 & 0.274 & 0.511 & 0.252 & 0.181 \\
\gr {Sparse4Dv2~\cite{sparse4dv2}$\ddagger$} & ResNet50 & 256$\times$704 & 53.9 & \red{43.9} & 0.598 & 0.270 & 0.475 & 0.282 & 0.179 \\
{StreamPETR~\cite{StreamPetr}$\ddagger$} & ResNet50 & 256$\times$704 & 54.0 & 43.2 & 0.581 & 0.272 & 0.413 & 0.295 & 0.195 \\
\gr {SparseBEV~\cite{SparseBEV}$\ddagger$} & ResNet50 & 256$\times$704 & \blue{54.5} & 43.2 & 0.606 & 0.274 & 0.387 & 0.251 & 0.186 \\
{\ours~(Ours)$\ddagger$} & ResNet50 & 256$\times$704 & \red{55.4} & \blue{43.7} & 0.512 & 0.262 & 0.367 & 0.285 & 0.213  \\
\hline

{PETR~\cite{PETR}*} & ResNet101-DCN & 512$\times$1408 & 44.1 & 36.6 & 0.717 & 0.267 & 0.412 & 0.834 & 0.190 \\
\gr {BEVDepth~\cite{BEVDepth}*$\dagger$} & ResNet101 & 512$\times$1408 & 53.5 & 41.2 & 0.565 & 0.266 & 0.358 & 0.331 & 0.190 \\
{BEVFormer~\cite{BEVFormer}*$\ddagger$} & ResNet101-DCN & 900$\times$1600 & 51.7 & 41.6 & 0.673 & 0.274 & 0.372 & 0.394 & 0.198 \\
\gr {HoP-BEVFormer~\cite{HoP}*$\ddagger$} & ResNet101-DCN & 512$\times$1408 & 55.8 & 45.4 & 0.565 & 0.265 & 0.327 & 0.337 & 0.194 \\
{StreamPETR~\cite{StreamPetr}*$\ddagger$} & V2-99 & 320$\times$800 & 57.1 & 48.2 & 0.569 & 0.262 & 0.315 & 0.257 & 0.199 \\
\gr {SOLOFusion~\cite{SOLOFusion}*$\ddagger$} & ResNet101 & 512$\times$1408 & 58.2 & 48.3 & 0.503 & 0.264 & 0.381 & 0.246 & 0.207 \\
{SparseBEV~\cite{SparseBEV}*$\ddagger$} & ResNet101 & 512$\times$1408 & 59.2 & 50.1 & 0.562 & 0.265 & 0.321 & 0.243 & 0.195\\
\gr {Far3D~\cite{far3d}*$\ddagger$} & ResNet101 & 512$\times$1408 & \blue{59.4} & \red{51.0} & 0.551 & 0.258 & 0.372 & 0.238 & 0.195 \\
{\ours~(Ours)*$\ddagger$} & V2-99 \& ResNet50 & 640$\times$1152 \& 256$\times$704 & \red{59.9} & \blue{50.2} & 0.465 & 0.261 & 0.335 & 0.267 & 0.197 \\

\hline
\end{tabular}
}
\end{table*}

\subsection{Main Results}

We compare the proposed \ours~with previous end-to-end multi-task models on the nuScenes \texttt{val} sets in~\tabref{table:exp1} and ~\figref{fig:multitask_result}.
\ours~shows favorable multi-task performance and achieves state-of-the-art results.
Specifically, \ours~outperforms BEVFormer~\cite{BEVFormer} by 7.9 NDS and 8.7 mAP on the 3D object detection task and 8.6 mIoU on the BEV semantic segmentation task.
As for PETRv2~\cite{PETRv2}, which shows excellent BEV semantic segmentation performance, our models surpass it by 10.4 NDS on the 3D object detection task while keeping competitive performance on the BEV semantic segmentation task.
%

In addition, we compare the training time per epoch (with the same batch size) with these methods.
HENet outperforms all other methods with less training time, showing the proposed HENet is more efficient.
Moreover, HENet can use more frames to improve performance, while other methods, like PETRv2, cannot use 9 frames due to the limitation of GPU memory.

\subsection{Analysis of Conflict in End-to-end Multi-task 3D Perception}
\label{sec:analysis}

To analyze the conflict between 3D object detection and BEV semantic segmentation tasks, we first conduct experiments on single tasks to find the optimal configuration of training and models, including the resolution of the input images, the number of temporal frames, and the BEV feature grid sizes.
As shown in the upper part of \figref{fig:diff}, we observe that the results of 3D object detection and BEV semantic segmentation tasks are consistently improved when the input resolutions and the number of temporal frames increase.
%
%
%
However, for BEV feature grid size, we find that different tasks prefer different BEV grid sizes.
We conjecture this inconsistency can be attributed to the tasks' characteristics.
The 3D object detection task focuses on localizing local foreground objects. In contrast, the BEV semantic segmentation task is required to comprehensively understand the large-scale scene, including lane lines and roads.
As shown in the bottom of \figref{fig:diff}, the BEV feature of detection is local highlighting, while the segmentation feature shows more activated regions globally. 
Therefore, when the BEV feature grid size is too small, the BEV semantic segmentation task suffers from aggregating local features and cannot capture global information for the whole scene understanding.
By contrast, larger BEV features can provide more local details for the 3D object detection task.

\begin{table*}[t]
\setlength{\belowcaptionskip}{-0.cm}
\centering
\caption[results3]{\textbf{Comparison of 3D object detection results on nuScences {\tt{test}} set.} The best and second best results are marked in \red{red} and \blue{blue}. 
\dag uses future frames. \ddag uses test-time augmentation.
}

\label{table:exp3} 
\resizebox{\textwidth}{!}{
\begin{tabular}{l|cc||cc|ccccc}
\hline
{\textbf{Methods}} & \textbf{Backbone} & \textbf{Input Size} & \textbf{NDS$\uparrow$} & \textbf{mAP$\uparrow$}  & \textbf{mATE$\downarrow$} & \textbf{mASE$\downarrow$} & \textbf{mAOE$\downarrow$} & \textbf{mAVE$\downarrow$} & \textbf{mAAE$\downarrow$}\\
\hline

{BEVDet4D~\cite{BEVDet4D}}  & Swin-B & 900$\times$1600 & 56.9 & 45.1 &0.511 & 0.241 & 0.386 & 0.301 & 0.121 \\
\gr {PolarFormer~\cite{polarformer}} & V2-99 & 900$\times$1600 & 57.2 & 49.3 & 0.556 & 0.256 & 0.364 & 0.439 & 0.127 \\
{PETRv2~\cite{PETRv2}}  & V2-99 & 640$\times$1600 & 58.2 & 49.0 & 0.561 & 0.243 & 0.361 & 0.343 & 0.120 \\
\gr {HoP-BEVFormer~\cite{HoP}} & V2-99 & 900$\times$1600 & 60.3 & 51.7 & 0.501 & 0.245 & 0.346 & 0.362 & 0.105 \\
{BEVDepth~\cite{BEVDepth}} & ConvNeXt-B & 640$\times$1600 & 60.9 & 52.0 & 0.445 & 0.243 & 0.352 & 0.347 & 0.127 \\
\gr {BEVStereo~\cite{BEVStereo}} & V2-99 & 640$\times$1600 & 61.0 & 52.5 & 0.431 & 0.246 & 0.358 & 0.357 & 0.138 \\
{SOLOFusion~\cite{SOLOFusion}} & ConvNeXt-B & 640$\times$1600 & 61.9 & 54.0 & 0.453 & 0.257 & 0.376 & 0.276 & 0.148 \\
\gr {AeDet~\cite{aedet}} & ConvNeXt-B & 640$\times$1600 & 62.0 & 53.1 & 0.439 & 0.247 & 0.344 & 0.292 & 0.130 \\
{BEVFormerv2~\cite{bevformerv2}} & InternImage-B & 640$\times$1600 & 62.0 & 54.0 & 0.488 & 0.251 & 0.335 & 0.302 & 0.122 \\
\gr {FB-BEV~\cite{FBBEV}} & V2-99 & 640$\times$1600 & 62.4 & 53.7 & 0.439 & 0.250 & 0.358 & 0.270 & 0.128 \\
{StreamPETR~\cite{StreamPetr}} & V2-99 & 640$\times$1600 & \blue{63.6} & 55.0 & 0.479 & 0.239 & 0.317 & 0.241 & 0.119 \\
\gr {SparseBEV~\cite{SparseBEV}} & V2-99 & 640$\times$1600 & \blue{63.6} & \blue{55.6} & 0.485 & 0.244 & 0.332 & 0.246 & 0.117\\
{Sparse4Dv2~\cite{sparse4dv2}} & V2-99 & 640$\times$1600 & \red{63.8} & \blue{55.6} & 0.462 & 0.238 & 0.328 & 0.264 & 0.115 \\
\gr {\ours~(Ours)} & V2-99 \& ResNet50 & 640$\times$1152 \& 256$\times$704 & \red{63.8} & \red{57.5} & 0.432 & 0.242 & 0.368 & 0.320 & 0.129 \\
\hline
\end{tabular}
}
\end{table*}

\subsection{Single Task Results}

Considering that many works on 3D perception only predict single-task results, we conduct experiments on single tasks and compare the results of \ours~with these task-specific models.
Through this comparison, we illustrate the superiority of our Hybrid Image Encoding Network and Temporal Feature Integration, and further demonstrate the effectiveness of \ours.

\noindent\textbf{3D Object Detection Results.} 
We present the results of \ours~for single 3D object detection task on the nuScenes \texttt{val} and \texttt{test} sets in~\tabref{table:exp2} and~\ref{table:exp3}, respectively.
%
As shown in ~\tabref{table:exp2}, \ours~surpasses all multi-view camera 3D object detection methods under different backbone and input resolution configurations, demonstrating the effectiveness of the proposed hybrid image encoding network and temporal feature integration module.
For online evaluation, \tabref{table:exp3} shows \ours~achieves state-of-the-art 3D object detection results.
%
%
%

\noindent\textbf{BEV Semantic Segmentation Results.} We present the results of \ours~for single BEV semantic segmentation task on the nuScenes \texttt{val} sets in~\tabref{table:exp4}. \ours~obtain competitive results compared to existing methods.

It is worth mentioning that compared with single-task performance, the end-to-end multi-task performance of \ours~ only drops 0.3 mAP and 0.8 mIOU for the 3D object detection task and the BEV semantic segmentation task, respectively. 
These results indicate the effectiveness of the independent BEV feature encoding design, as well as the use of different BEV grid sizes for addressing multi-task conflict problems.

\begin{table}[t]
\setlength{\belowcaptionskip}{-0.cm}
\centering
\caption[results3]{\textbf{Comparison of BEV semantic segmentation results on nuScences {\tt{val}} set.} The best and second best results are marked in \red{red} and \blue{blue}.}

\label{table:exp4} 
\resizebox{0.8\columnwidth}{!}{
\begin{tabular}{l|c|c|ccc}
\hline
{\textbf{Methods}} & \textbf{Backbone} & $\textbf{mIoU}$ & $\textbf{mIoU}_{\textbf{veh}}\uparrow$ & $\textbf{mIoU}_{\textbf{area}}\uparrow$ & $\textbf{mIoU}_{\textbf{div}}\uparrow$\\
\hline

\gr {VPN~\cite{VPN}} & ResNet50 & 42.7 & 31.8 & 76.9 & 19.4 \\
{LSS~\cite{LSS}} & ResNet50 & 46.5 & 41.7 & 77.7 & 20.0 \\
\gr {BEVFormer~\cite{BEVFormer}} & ResNet101-DCN & 50.2 & 44.8 & 80.1 & 25.7 \\
{FIERY~\cite{fiery}} & ResNet-101 & - & 38.2 & - & - \\
\gr {M2BEV~\cite{xie2022m}} & ResNeXt-101 & - & - & 77.2 & 40.5 \\
{PETRv2~\cite{PETRv2}} & V2-99 & \red{60.3} & \blue{46.3} & \red{85.6} & \red{49.0} \\

\gr {\ours~(Ours)} & V2-99 \& ResNet50 & \blue{58.8} & \red{51.6} & \blue{82.3} & \blue{42.4} \\

\hline
\end{tabular}
}
\end{table}

\begin{table*}[t]
\setlength{\belowcaptionskip}{-0.cm}
\centering
\caption{\textbf{Ablation of Hybrid Image Encoding Network on Detection Task.} 
`Time' denotes the overall training time.
`BS' indicates the batch size.
FPS is measured on a single RTX3090 with FP32. Training cost (GPU memory and time) is estimated on 8$\times$Tesla A800 GPUs with FP32.}

\label{table:abl1} 
\resizebox{\textwidth}{!}{
\begin{tabular}{c|l|ccc||cc|ccc}
\hline

& \textbf{Model} & \textbf{Backbone} & \textbf{Input} & \textbf{Frames} & \textbf{NDS} & \textbf{mAP} & \textbf{FPS} & \textbf{GPU memory} & \textbf{Time}\\
\hline
\gr A & BEVDepth4D & R50 & 256$\times$704 & 9 frames & 53.8 & 40.9 & 19.1 & 14.1G / BS=16 & 16h \\
B & BEVStereo & V2-99 & 640$\times$1152 & 2 frames & 58.2 & 48.0 & 4.52 & 37.0G / BS=16 & 48h \\
\gr C & BEVStereo & V2-99 & 896$\times$1600 & 2 frames & 59.2 & 50.0 & 2.69 & 68.4G / BS=16 & 96h \\
A+B & Model Ensemble &-&-&- & 57.3 & 46.4 & 3.53 & - & - \\
\gr A+B & HENet w/o AFFM & V2-99 \& R50 & 640$\times$1152 \& 256$\times$704 & 2 \& 7 frames & 59.2 & 49.9 & 3.71 & 42.3G / BS=16  & 27h \\
A+B & \ours & V2-99 \& R50 & 640$\times$1152 \& 256$\times$704 & 2 \& 7 frames & 59.9 & 50.2 & 3.65 & 43.5G / BS=16 & 27h\\




\hline

\gr D & BEVDepth4D & R18 & 640$\times$1152 & 2 frames & 48.6 & 34.8 & 14.3 & 31.2G / BS=64 & 14h \\
E & BEVDepth4D & R18 & 256$\times$704 & 9 frames & 48.6 & 34.9 & 21.7 & 22.9G / BS=64 & 14h \\
\gr F & BEVDepth4D & R18 & 640$\times$1152 & 9 frames & 51.2 & 39.2 & 14.1 & 43.5G / BS=64 & 44h \\
D+E & Model Ensemble &-&-&- & 48.9 & 35.2 & 7.93 & - & - \\
\gr D+E & \ours & R18 \& R18 & 640$\times$1152 \& 256$\times$704 & 2 \& 7 frames & 52.1 & 39.8 & 8.91 & 41.3G / BS=64 & 13h \\



\hline
\end{tabular}
}
\end{table*}

\subsection{Ablation}

We also conduct ablation studies for each proposed module on nuScenes \texttt{val} set.
In the following experiment, we adopt BEVDepth4D~\cite{BEVDepth} as the small image encoder and BEVStereo~\cite{BEVStereo} as the large image encoder.

\begin{table*}[t]
\setlength{\belowcaptionskip}{-0.cm}
\centering

\caption{\textbf{Ablation of Hybrid Image Encoding Network on Multi-task.} }

\label{table:abl1b} 
\resizebox{\columnwidth}{!}{
\begin{tabular}{c|l|ccc|cc}

\hline
& \textbf{Model} & \textbf{Backbone} & \textbf{Input} & \textbf{Frames} & \textbf{NDS} & \textbf{mIoU}\\
\hline
\gr A & BEVDepth4D & R50 & 256$\times$704 & 9 frames & 53.0 & 51.5 \\
B & BEVStereo & V2-99 & 640$\times$1152 & 2 frames & 58.0 & 55.8\\
\gr C & BEVStereo & V2-99 & 896$\times$1600 & 2 frames & 58.9 & 56.7\\
A+B & HENet w/o AFFM & V2-99 \& R50 & 640$\times$1152 \& 256$\times$704 & 2 \& 7 frames & 59.0 & 56.9\\
\gr A+B & HENet w/ AFFM & V2-99 \& R50 & 640$\times$1152 \& 256$\times$704 & 2 \& 7 frames & 59.9 & 58.0\\

\hline
\end{tabular}
}
\end{table*}

\begin{figure*}[t]
    \setlength{\abovecaptionskip}{-0.cm}
    \centering
    \includegraphics[width=\linewidth]{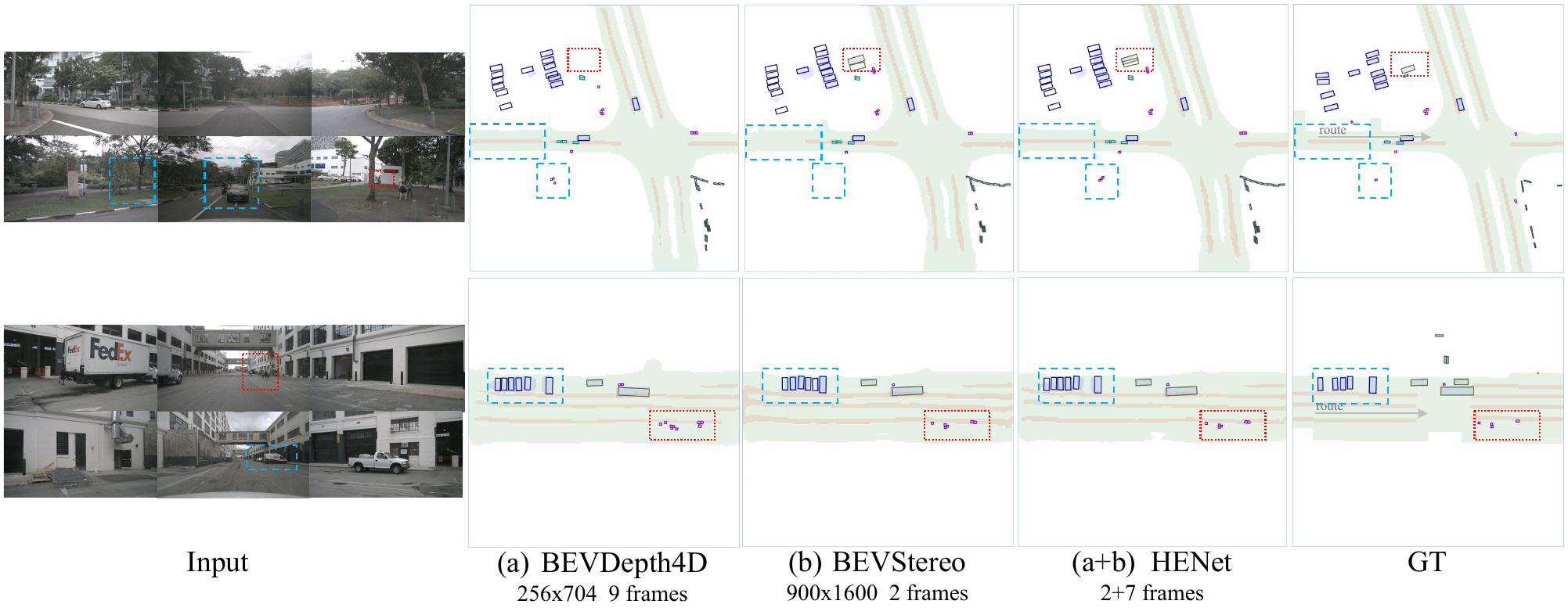}
    \caption{\textbf{Visualization results of \ours~and baselines on end-to-end multi-tasking.}
    From left to right, we show multi-view image inputs, results of BEVDepth4D, BEVStereo, and \ours~(BEVDepth4D + BEVStereo), and the ground truth. 
    The proposed \ours~ estimates occluded objects better through long-term information and has more accurate predictions through high-resolution information.
    }
    \label{fig:vis}

\end{figure*}

\setlength{\tabcolsep}{0.4em}
\begin{table}[t]
\setlength{\belowcaptionskip}{-0.cm}
\centering
\caption{\textbf{Ablation of Temporal Feature Integration module.}
Our proposed backward and forward processes with AFFM achieve the best results.
}

\definecolor{dr}{HTML}{ea4335}
\label{table:abl2} 
\resizebox{\columnwidth}{!}{
\begin{tabular}{l|cc|c}
\hline



Temporal Integration & NDS & mAP & Parameters \\
\hline
Global Concatenation\&Conv (BEVDepth4D~\cite{BEVDet4D}) & 52.3  & 40.8 & 76.51M \\
\gr Global Concatenation\&Conv + larger BEV encoder  & 52.4  & 40.7 & 77.70M \\
Global attention & 52.6 & 40.9 & 76.68M \\
\gr Forward with adjacent Concatenation\&Conv & 52.8  & 40.6 & 76.63M \\
Forward with AFFM & 53.1  & 41.2 & 76.64M \\
\gr Backward + Forward with adjacent Concatenation\&Conv & 52.8  & 40.7 & 76.63M \\
Backward + Forward with AFFM (Ours) & 53.2  & 41.5 & 76.64M \\

\hline
\end{tabular}

}
\end{table}

\setlength{\tabcolsep}{0.3em}
\begin{table}[t]
\setlength{\belowcaptionskip}{-0.cm}
\centering
\caption{\textbf{Ablation of Independent BEV Feature Encoding} `AFS' is the adaptive feature selection. `IE' denotes the independent BEV encoder. All experiments only used a single BEVDepth4D with ResNet-50 as the image encoder.}

\definecolor{dr}{HTML}{ea4335}
\label{table:abl3} 
\resizebox{0.7\columnwidth}{!}{
\begin{tabular}{cc|cc||cc|c}
\hline

\textbf{Det-grid} & \textbf{Seg-grid} & \textbf{AFS} & \textbf{IE} & \textbf{NDS} & \textbf{mAP} & $\textbf{mIoU}$ \\

\hline

0.4m & 0.4m  &  &  & 53.2 & 41.9 & 41.6 \\
0.8m & 0.8m  &  &  & 50.5 & 39.6 & 50.9 \\
0.4m & 0.8m &  &  & 52.9 & 42.0 & 50.6 \\
0.4m & 0.8m & \checkmark &  & 53.3 \textcolor{dr}{$\uparrow$0.4} & 42.3 \textcolor{dr}{$\uparrow$0.3} & 51.2 \textcolor{dr}{$\uparrow$0.6} \\
0.4m & 0.8m & \checkmark & \checkmark & 54.6 \textcolor{dr}{$\uparrow$1.7} & 43.1 \textcolor{dr}{$\uparrow$1.1} & 54.0 \textcolor{dr}{$\uparrow$3.4} \\

\hline
\end{tabular}

}
\end{table}

\noindent\textbf{Hybrid Image Encoding Network}.
To demonstrate the effectiveness of the proposed hybrid image encoding network, we compare \ours~with the two baseline methods and their ensemble model.
%
As shown in \tabref{table:abl1}, by combining with BEVDepth4D~\cite{BEVDepth} and BEVStereo~\cite{BEVStereo} through hybrid image encoding, \ours~can significantly improve the 3D object detection performance. 
Compared to increasing resolution (model C), Hybrid Image Encoding Network can achieve higher accuracy with faster inference speed and lower training costs.
Compared to increasing the frame number (model F), Hybrid Image Encoding Network can achieve higher accuracy with lower training costs.
Notably, ensembling the results of the two baselines by NMS decreases the overall performance since the weaker BEVDepth4D~\cite{BEVDepth} introduces many false positive detection results.

For better comparison, we provide a multi-task ablation study, as shown in \tabref{table:abl1b}.

We also provide the visualization of the detection results in \figref{fig:vis}. 
It can be seen that, due to motion or occlusion, some objects or scenes (as shown in the blue boxes) require a longer time sequence. 
Besides, high-resolution and sophisticated depth estimation methods benefit the perception of difficult objects and scenes (as shown in the red boxes).
\ours~can effectively combine the advantages of long-time sequence, high-resolution, and sophisticated depth estimation.


\noindent\textbf{Temporal Feature Integration.}
\tabref{table:abl2} compares the results between different types of temporal feature integration methods.
%
%
%

Our adjacent attention achieves the best results.
%
We observe that adjacent design is more effective than global operation, whether using attention or using Concatenation\&Conv.
%
%
Besides, compared to concatenation and convolution, our AFFM, which is based on the attention mechanism, performs better.
Lastly, global attention and the larger BEV encoder introduce more model parameters and achieve worse performance than pair-wise attention. This demonstrates that the performance improvements in pair-wise attention come from the design itself rather than the increased model parameters.

\noindent\textbf{Independent BEV Feature Encoding.}
As analyzed in Sec.\ref{sec:analysis}, 3D object detection and BEV semantic segmentation tasks prefer different BEV feature grid sizes.
%
%
As shown in \tabref{table:abl3}, experimental results show that using BEV feature maps in different sizes for different tasks achieves the best multi-task performance trade-off.
Moreover, adopting independent adaptive feature selection and BEV encoder for each task can further improve the multi-task performance of 1.7 NDS, 1.1 mAP, and 3.4 mIoU.

%% file: sec/5_conclusion.tex
\section{Conclusion}
\label{sec:conclusion}

In this paper, we present \ours, an end-to-end framework for multi-task 3D perception. 
We propose a hybrid image encoding network and the temporal feature integration module to deal efficiently with high-resolution and long-term temporal image inputs.
%
Besides, we adopt task-specific BEV grid sizes, independent BEV feature encoder and decoder to address multi-task conflict issue.
%
%
Experimental results show that \ours~obtains the state-of-the-art multi-task results on nuScenes, including 3D object detection and BEV semantic segmentation.